\documentclass[runningheads]{llncs}
\usepackage{amsmath,amssymb,array,graphicx,color,ulem}
\newcommand\norm[1]{\left\lVert#1\right\rVert}

\usepackage{xcolor}
\usepackage[colorlinks = true,
            linkcolor = blue,
            urlcolor  = blue,
            citecolor = blue,
            anchorcolor = blue]{hyperref}

\begin{document}
\title{
Semi-supervised Medical Image Classification 
with Global Latent Mixing}
\author{Prashnna Kumar Gyawali, Sandesh Ghimire, Pradeep Bajracharya, Zhiyuan Li \and Linwei Wang}
\institute{\email{pkg2182@rit.edu} \\ Rochester Institute of Technology}
\authorrunning{Gyawali et al.}
\maketitle

\begin{abstract}
Computer-aided diagnosis via deep learning relies on large-scale annotated data sets, 
which can be costly when involving expert knowledge. Semi-supervised learning (SSL) mitigates this challenge by leveraging unlabeled data.
One effective SSL approach 
is to regularize the local smoothness of neural functions 
via perturbations around single data points. 
In this work, we argue that 
regularizing the global smoothness of neural functions 
by filling the void in between data points 
can further improve SSL. 
We present a novel SSL approach 
that trains the neural network 
on linear mixing of labeled and unlabeled data, 
at both the input and latent space 
in order to 
regularize different portions of the network. 
We evaluated the presented model on two distinct medical image data sets for semi-supervised classification of thoracic disease
and skin lesion, 
demonstrating its improved performance over 
SSL with local perturbations 
and SSL with global mixing but at the input space only.
Our code is available at \href{https://github.com/Prasanna1991/LatentMixing}{https://github.com/Prasanna1991/LatentMixing}.
\vspace{-0.2cm}
\end{abstract}

\section{Introduction}
Medical image analysis 
via deep learning has achieved strong performance when 
supervised with a large labeled data set. 
Collecting such data sets 
is however costly 
in the medical domain 
since it involves expert knowledge. Semi-supervised learning (SSL) 
mitigates this challenge 
by leveraging unlabeled data.

An important goal in SSL 
is to avoid over-fitting the network function 
to small labeled data. 
A common inductive bias 
to guide 
this 
is 
the assumption of \textit{smoothness} or \textit{consistency} of the network function, \textit{i.e.,} nearby points and points of the same manifold 
should have the same label predictions.
For instance, self-ensembling \cite{laine2016temporal} penalizes inconsistent predictions of unlabeled data under local perturbations, 
and virtual adversarial training \cite{miyato2018virtual} maintains consistency by forcing predictions of different adversarially-perturbed inputs to be the same. 

By considering perturbations around single data points, 
these approaches regularize 
only the \textit{local} smoothness of the network function
in the vicinity of available 
data points: 
no constraint is imposed on the global behavior of the network function 
in between data points 
\cite{luo2018smooth}. 
To better exploit the structure of unlabeled data, 
we consider a strategy of \textit{mixup} which was recently proposed to 
train a deep network on a linear mixing of 
pairs of input data and their corresponding labels \cite{zhang2017mixup}.
By filling the void between input samples, 
this strategy 
regularizes the \textit{global} smoothness of the function  
and was shown to
improve the generalization of state-of-the-art neural architectures 
in both supervised \cite{zhang2017mixup} and semi-supervised learning \cite{berthelot2019mixmatch}.
This mixup strategy was recently extended to 
the latent space, 
showing improvement over mixing in the input space only,
in a supervised setting. 

We argue that the mixup strategy -- training a network on linear mixing of 
input data and their labels --  
can be interpreted as 
regularizing the network 
to
approximate a linear interpolation function 
in between data points. 
The gain of performance brought by 
mixing in the latent space, 
therefore, 
is partly owing to relaxing this 
linearity constraint
to 
a portion of the network between 
the selected latent space 
and the output space.
We also hypothesize that,  
since high-level representations in deep-networks 
encode important information for discriminative tasks, 
mixing at the latent space may provide novel training signals for SSL. 

Therefore, 
we propose to extend this regularization, 
\textit{i.e.}, regularizing different portions of the network 
between the latent space and output space, 
for SSL and 
demonstrate its first application in medical image classification. 
In this approach, 
we perform linear mixing of pairs of labeled and unlabeled data 
-- both in the input and latent space -- along with 
their corresponding labels: 
for the latter, the label is \textit{guessed} and continuously updated from
an average of predictions of augmented samples 
for each unlabeled data point.
We evaluate the presented SSL model on 
two distinct medical image classification tasks: 
multi-label classification of thoracic disease 
using Chexpert lung X-ray images \cite{irvin2019chexpert}, 
and skin disease classification using Skin Lesion images \cite{codella2019skin,tschandl2018ham10000}. 
We compare the performance of the presented method 
with both a supervised baseline, 
and several SSL methods 
including mixup at the input space \cite{zhang2017mixup}, 
standard self-ensembling 
in the input space
\cite{laine2016temporal}, 
and recently-introduced self-ensembling 
at the latent space \cite{gyawali2019semi}. 
We further provide ablation studies 
and analyze the effect of 
function smoothing achieved by the presented method. 

\section{Related Work}
\subsubsection{SSL in Medical Image Analysis:}
Many recent semi-supervised works in medical image analysis have focused on explicitly regularizing the local smoothness of the neural function \cite{bortsova2019semi,peng2020deep,gyawali2019semi}.
For instance, in \cite{bortsova2019semi}, 
a siamese architecture for both labeled and unlabeled data points was proposed to encourage consistent segmentation under a given class of transformations. 
In \cite{peng2020deep}, ensemble diversity was enforced with the use of adversarial samples to improve 
semi-supervised 
semantic image segmentation. 
In \cite{gyawali2019semi}, the disentangled stochastic latent space was learned to improve self-ensembling for
semi-supervised classification 
of chest X-ray images.
In these works, each data point was subjected to local perturbations, \textit{e.g.}, elastic deformations \cite{bortsova2019semi}, virtual adversarial direction \cite{peng2020deep}, or sampling from latent posterior distributions \cite{gyawali2019semi}, for \textit{local} smoothness regularization. 

In \cite{luo2018smooth}, 
the idea of promoting \textit{global} smoothness in SSL was explored 
by constructing a teacher graph network. 
Similar approaches 
exploiting 
the global smoothness of neural functions, however, has not been studied in medical images.

\subsubsection{Regularization with the Mixup Strategy: }
The mixup strategy 
was first presented in \cite{zhang2017mixup} 
to improve generalization of supervised models
by mixing the data pairs at the input space. 
It was recently extended in a 
semi-supervised 
setting where 
the mixing is considered for both labeled and unlabeled data points \cite{berthelot2019mixmatch}.
In the meantime, 
a similar idea was also extended 
to the mixing of 
hidden representations \cite{verma2019manifold}, 
demonstrating improvement over mixing at the input space, 
although only in supervised learning. 

To our knowledge, this is the first semi-supervised classification network 
that employs the mixup strategy at the latent space, 
and the first time this type of approaches is applied to 
semi-supervised medical image classification. 

\begin{figure*}[tb]
\begin{center}
\includegraphics[width=11cm,height=11cm,keepaspectratio]{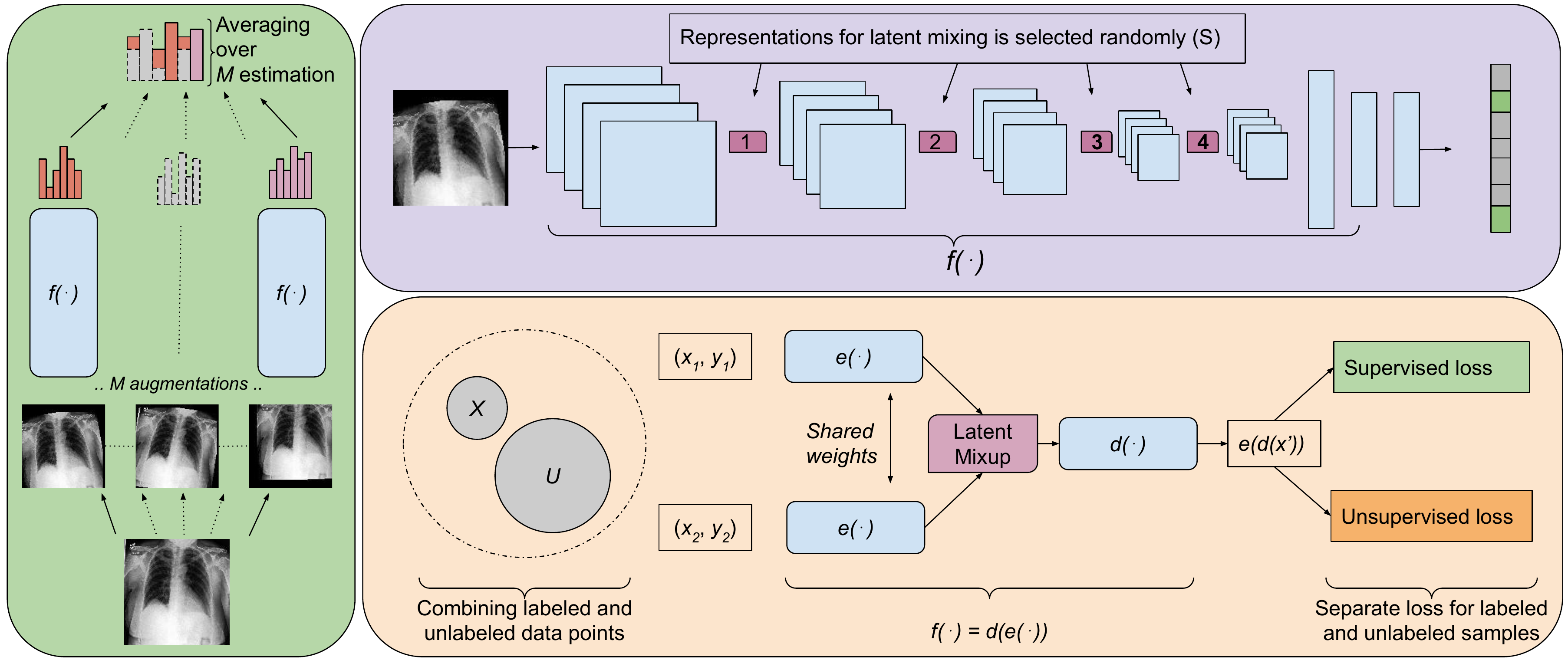}
\end{center}
\vspace{-0.4cm}
\caption{Schematic diagram of the presented SSL method. 
During training, 
we continuously guess labels for the unlabeled data points (left) 
and then perform SSL via mixing at the input and latent space (bottom right). On the top right, we demonstrate the layers in the deep network where latent representations can be mixed.\vspace{-.3cm}} 
\label{fig:modelDiagram}
\end{figure*}

\vspace{-.3cm}

\section{Methodology}
\vspace{-.2cm}

We consider a set of labeled training examples $\mathcal{X}$ with the corresponding labels $\mathcal{Y}$, and a set of unlabeled training samples $\mathcal{U}$. We aim to learn parameters $\theta$ for the mapping function $f: \mathcal{X} \rightarrow\mathcal{Y}$, approximated via a deep neural network. 
Along the course of the training, 
we first \textit{guess} and 
continuously update the labels for unlabeled data points 
(section \ref{sec:label}).
We then perform linear mixing between 
labeled and unlabeled data points, 
both in the input and latent space, 
along with their corresponding actual or guessed labels 
(section \ref{sec:mixup}). 
Finally, 
the SSL model is trained on the mixed data sets using 
different losses depending on 
whether the mixed data point is closer to
labeled or unlabeled data (section \ref{sec:loss}).
Fig. \ref{fig:modelDiagram} summarizes 
the key components of this semi-supervised learning process.

\subsection{Guessing Labels}
\label{sec:label}
We 
guess the labels for unlabeled data by augmenting $M$ separate copies of 
data batch $u_b$, 
and computing the average of the model's prediction as:
\begin{equation}
  \label{eq:labelEstimation}
q_b = \frac{1}{M}\sum_{m=1}^{M} f(u_{b, m}; \theta)
\end{equation}
The label guessing in this manner implicitly works as consistency regularization as the input transformations are assumed to leave class semantics unaffected. 
The guessed labels are continuously changed 
as the neural function $f(\mathbf{x})$ is updated over the course of the training. 

\subsection{Input and Latent Mixup}
\label{sec:mixup}
Since the mapping function $f(\mathbf{x})$ is approximated by deep neural network, we can decompose this function as $f(\mathbf{x})$ = $d_{l}(e_{l}(\mathbf{x}))$, where $e_{l}$ represents the part of the neural network that encodes the input data to some latent representation at layer $l$, and $d_{l}$ denotes the part of neural network that decodes such latent representation to the output $f(\mathbf{x})$. 
Inspired by \cite{verma2019manifold}, we determine a set of eligible layers $\mathcal{S}$ in the neural network from which we randomly select a layer $l$ and apply mixup in that layer (schematics in top-right; Fig. \ref{fig:modelDiagram}).
For each batch, we combine and shuffle labeled and unlabeled data points 
to obtain a pair of random mini-batches ($\mathbf{x}_{1}$, $\mathbf{y}_{1}$) and ($\mathbf{x}_{2}$, $\mathbf{y}_{2}$). We pass these pairs to $e_{l}$ to obtain latent pairs ($e_{l}(\mathbf{x}_{1})$, $\mathbf{y}_{1}$) and ($e_{l}(\mathbf{x}_{2})$, $\mathbf{y}_{2}$), 
and then perform mixup at this latent layer 
to
produce the mixed minibatch as ($e_{l}(\mathbf{x})'$, $\mathbf{y}'$) as:
\begin{equation}
  \label{eq:mixup}
  \begin{aligned}
  \lambda & \sim \text{Beta}(\alpha, \alpha)\\
  \lambda' & = \text{max}(\lambda, 1-\lambda) \\
  e_{l}(\mathbf{x})' & = \lambda' \cdot e_{l}(\mathbf{x}_{1}) + (1 - \lambda') \cdot e_{l}(\mathbf{x}_{2}) \\
  \mathbf{y}' & = \lambda' \cdot \mathbf{y}_{1} + (1 - \lambda') \cdot \mathbf{y}_{1}
  \end{aligned}
\end{equation} 
where $\alpha$ is the positive shape parameter of the Beta distribution, treated as hyperparameter in this work.
Because the mixing could occur between labeled and unlabeled data, 
it is important to ensure that the mixed data 
fairly represent the distribution of both labeled and unlabeled data. 
Furthermore, 
as will be described in section 
\ref{sec:loss}, 
different losses will be used 
to reflect a different treatment of 
the actual and guess labels 
due to their difference in reliability. 
It is thus also important to know  
whether each mixed data point  
is closer to labeled 
or unlabeled data. 
To do so,
we use $\lambda'$ instead of $\lambda$ in equations (\ref{eq:mixup}) 
to ensure that $e_{l}(\mathbf{x})'$ 
is always closer to $e_{l}(\mathbf{x}_{1})$ 
than to $e_{l}(\mathbf{x}_{2})$, 
allowing us to 
rely on the knowledge of $\mathbf{x}_{1}$ 
to determine which loss to apply on the mixed data point.


Depending upon $\mathcal{S}$, we achieve different mixup strategies. For example,  when $\mathcal{S} = \{0\}$, we only mix at the input space. 
When $\mathcal{S} = \{0, 1\}$,
we mix at the input space and latent layer 1. 
When $\mathcal{S} = \{1\}$, we mix only at the latent layer 1. 

\subsection{Supervised and Unsupervised Loss}
\label{sec:loss}
To treat the actual and guessed labels differently 
because the latter are less reliable, 
we use different losses 
for data points that are 
closer to labeled versus unlabeled data. 
For data points in a batch $\mathcal{B}$  
that are closer to labeled data, 
the loss term $\mathcal{L}_{\mathcal{X}}$ 
is the cross-entropy loss: 
\begin{equation}
  \label{eq:xloss}
  \begin{aligned}
\mathcal{L}_{\mathcal{X}} = 
\underset{(\mathcal{B}\cap\mathcal{X})}{\sum} \underset{l \sim \mathcal{S}}{\sum} \ell (d_{l}(e_{l}(\mathbf{x})'), \mathbf{y}')
  \end{aligned}
\end{equation} 
For data points in $\mathcal{B}$  
that are closer to unlabeled data, 
the loss function 
is defined as a $L_{2}$ loss 
because it is considered to be less sensitive to 
incorrect predictions: 
\begin{equation}
  \label{eq:uloss}
  \begin{aligned}
\mathcal{L}_{\mathcal{U}} = 
\underset{(\mathcal{B}\cap\mathcal{U})}{\sum} \underset{l \sim \mathcal{S}}{\sum}  \norm{d_{l}(e_{l}(\mathbf{x})') - \mathbf{y}'}_{2}^{2} 
  \end{aligned}
\end{equation} 

After obtaining \textit{mixed} latent representation,
the network is optimized by minimizing the sum of these two losses: 
\begin{equation}
  \label{eq:loss}
  \begin{aligned}
\mathcal{L} = \mathcal{L}_{\mathcal{X}} + \lambda_{\mathcal{U}} \cdot \mathcal{L}_{\mathcal{U}}
  \end{aligned}
\end{equation} 
where $\lambda_{\mathcal{U}}$ is the weight term for the unsupervised loss.

    


\section{Experiments}
We first test the effectiveness of the presented SSL approach on 
two distinct benchmark data sets 
for 
medical image classifications, in comparison to 
a supervised baseline and alternative SSL models. 
We then analyze the effect of mixing at different latent layers, 
and 
perform ablation studies to assess 
the impact of different hyperparameters 
and the depth of latent mixing 
on 
the presented method. 
Finally, we discuss the effect 
of function smoothing achieved by the presented SSL strategy.

\subsection{Data sets}
\label{sec:dataset}
We evaluate the presented model on two open-sourced large-scale medical dataset: Chexpert \cite{irvin2019chexpert} and ISIC 2018 Skin Lesion Analysis \cite{codella2019skin,tschandl2018ham10000}. 

\subsubsection{Chexpert X-ray image classification:} 
Chexpert comprises of 224316 chest radiograph images from more than 60000 patients with labels for 14 different pathology categories. 
For pre-processing, we
removed all uncertain and lateral-view samples from the data set, 
and re-sized the images to 128x128 in dimension. 
To ensure a fair comparison, we used the publicly available data splits for the labeled training set (ranging from 100 to 500 samples), 
unlabeled set, 
validation set, 
and test set \cite{gyawali2019semi}. 
For data augmentation, we rotated an image in the range of (-10$^o$, 10$^o$) and shifted (horizontal and vertical) it in the range of (0, 0.1) fraction of the image. 
\vspace{-0.3cm}

\subsubsection{Skin image classification:}
ISIC 2018 skin data set comprises of 10015 dermoscopic images with labels for seven different disease categories.
Three sets of labeled training data (350, 600, and 1200) 
were created considering class balance. 
The same data re-sizing and data augmentation strategies as applied to X-ray images 
were applied here.

\subsection{Implementation details}
In our experiments, we use the AlexNet-inspired network from \cite{gyawali2019semi} to match their model implementation 
and training procedure closely. The network consists of five convolution blocks, followed by three fully-connected layers. 
All the models were trained up to 256 epochs 
with a learning rate of 1e-4 
and decayed 
by a factor of 10 at the 50th and 125th epochs. 
For 
label guessing, we used $M$ = 2 copies of unlabeled data.  
For Chexpert, unless mentioned otherwise, we used a set of eligible layers $\mathcal{S}$ = \{0, 2, 4\}, mixing parameter $\alpha$ = 1.0 for input mixup and $\alpha$ = 2.0 for latent mixup, 
and $\lambda_{\mathcal{U}}$ = 75 for the weight on unsupervised loss. 
For the skin data set, we used a set of eligible layers $\mathcal{S}$ = \{0, 1\}, mixing parameter $\alpha$ = 1.0 for both input and latent mixup, 
and $\lambda_{\mathcal{U}}$ = 50 for the weight on unsupervised loss. 
We used the separately held out validation set to determine the best model 
along the course of the training, 
and report the results on the test set. The code used in the experiments will be made publicly available.

\begin{table}[t]
\centering
    \begin{tabular}[t]{|l||c|c|c|c|c||c|c|c|}
    \hline
    Model & \multicolumn{5}{c||}{Chexpert ($k$)} & \multicolumn{3}{c|}{Skin ($k$)}\\ \cline{2-9}
    & 100 & 200 & 300 & 400 & 500 & 350 & 600 & 1200\\
    \hline
    Supervised baseline & 0.5576 & 0.6166 & 0.6208 & 0.6343 & 0.6353 & 0.7707 & 0.7991 & 0.8538\\
    Input Mixup & 0.6491 & 0.6627 & 0.6731 & 0.6779 & 0.6823 & 0.8504 & 0.8609 & 0.9040 \\
    Latent Mixup & \textbf{0.6523} & 0.6632 & \textbf{0.6747} & 0.6795 & 0.6836 & 0.8536 & 0.8736 & 0.9036\\
    Input+Latent Mixup & 0.6512 & \textbf{0.6641} & 0.6739 & \textbf{0.6796} & \textbf{0.6847} & \textbf{0.8666} & \textbf{0.8768} & \textbf{0.9073}\\
    \hline
    \end{tabular} \vspace{.1cm}
    \caption{
    Mean AUROC of 14 categories in the Chexpert data and seven categories in the skin data. 
    The reported values are the average of five random seeds runs. 
    }
\label{tab:classification_table1}
\vspace{-0.5cm}
\end{table}

\begin{table}[t]
\centering
    \begin{tabular}[t]{|l|c|c|c|c|c|}
    \hline
    Model & \multicolumn{5}{c|}{Chexpert ($k$)} \\ \cline{2-6}
    & 100 & 200 & 300 & 400 & 500\\
    \hline
    Image-space self-ensembling (noise) & 0.6012  & 0.6277  & 0.6444  & 0.6550 & 0.6626 \\
    Image-space self-ensembling (augmentation) & 0.6089  & 0.6301  & 0.6423  & 0.6530 & 0.6617 \\
    Latent-space self-ensembling  & 0.6200 & 0.6386 & 0.6484 & 0.6637 & 0.6697 \\
    \textbf{Input + Latent Mixup} \textit{(ours)} & \textbf{0.6512} & \textbf{0.6641} & \textbf{0.6739} & \textbf{0.6796} & \textbf{0.6847} \\
    \hline
    \end{tabular} \vspace{.1cm}
    \caption{Mean AUROC for classification for 14 categories in the Chexpert data. The average of five randomly-seeded runs is reported by the presented method, whereas the best result is reported for the other method based on \cite{gyawali2019semi}.
    }
\label{tab:classification_table}
\vspace{-0.4cm}
\end{table}

\subsection{Results} 

\subsubsection{Comparison studies:} 
In both data sets, 
we first evaluate the SSL performance of the presented model 
in comparison with two baselines: a fully-supervised baseline where we train the network with a supervised cross-entropy loss without mixing, 
and input mixup where SSL is performed 
with mixing at the input space only. 
The results are presented in Table \ref{tab:classification_table1}. 
For the presented approach, 
we present two versions: 
mixing only at the latent space (latent mixup SSL), 
and combining both input and latent mixing (input + latent mixup SSL).
As shown, 
mixing in the latent space 
in general improved the SSL performance 
over the baseline methods. 
Among the alternatives involving 
latent mixup, 
combined input and latent mixing yielded the best performance in three out of five cases in the Chexpert data set, and in all cases in the skin dataset. 

Using the Chexpert data set, 
we further compared the presented model 
with existing SSL methods 
that focused on regularizing 
\textit{local} smoothness of the network function 
via perturbations around single data points: 
self-ensembling at the input space \cite{laine2016temporal} 
using Gaussian noise perturbations (with std=0.15, image-space self-ensembling (noise)) 
or augmention with random translation and rotation 
(image-space self-ensembling (augmentation)), 
and ensembling at the disentangling latent space 
(latent-space self-ensembling) \cite{gyawali2019semi}. 
The results, as presented in \ref{tab:classification_table}, 
showed a clear improvement of the presented method, 
supporting the advantage of 
regularizing the global in addition to local smoothness 
of neural functions. 

\begin{table}[t]
\vspace{-0.4cm}
    \begin{minipage}{.50\linewidth}
      \centering
   \begin{tabular}[t]{l|c|c}
    \hline
    Ablation & Latent mixup & Input + Latent \\
    ($K$ = 300) & &mixup\\
    \hline
    Presented & 0.6747 $\pm$ 0.23 & 0.6739 $\pm$ 0.20 \\
    Noise & 0.6508 $\pm$ 0.13 & 0.6512 $\pm$ 0.06 \\
    $\alpha$ = 1.0 & 0.6736 $\pm$ 0.17 & 0.6743 $\pm$ 0.11 \\
    $\lambda_{\mathcal{U}} = 100$ & 0.6722 $\pm$ 0.10 & 0.6719 $\pm$ 0.20 \\
    \hline
    \end{tabular} 
    \end{minipage} 
    \begin{minipage}{.50\linewidth}
        \includegraphics[width = \textwidth]{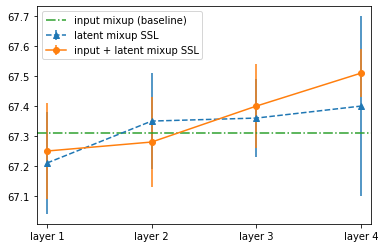}
    \end{minipage}%
    \caption{Effect of hyperparameters (left) and the latent depth for mixing (right). }
    \label{tab:ablation}
    \vspace{-0.3cm}
\end{table}
\subsubsection{Ablation studies:}
\label{sec:ablation}
We study the effect of different hyperparameters and elements in the presented SSL method, 
using a labeled dataset of size 300. 
The results are shown in Table \ref{tab:ablation} (left). 
While each had certain effect on the model performance, 
the most notable difference 
came from the data augmentation strategy used in the presented SSL method: 
replacing the presented data augmentation 
with image-level noises notably reduced the model performance, 
although still at a level higher 
than the ensembling baselines presented in Table \ref{tab:classification_table}. 

In Table \ref{tab:ablation} (right), we 
show how the model performance was affected 
by the depth of latent space 
at which the mixing was performed, 
in comparison to a fixed baseline (green dashed) 
of mixing at the input space only. 
As shown, 
mixing at the deeper layers of the network 
appeared to be more beneficial in general. 
This implies that 
it may be more appropriate to 
apply the linearity constraint, 
considering its limited function capacity, 
to the later portion of a deep neural network. 
It may also suggest that, 
since higher-level representations are more task-related, 
mixing in such space could help in generalization.  

\subsection{The effect of function smoothing}
\label{sec:smoothing} 

\begin{figure}[t]
\begin{center}
\includegraphics[width=10cm,height=10cm,keepaspectratio]{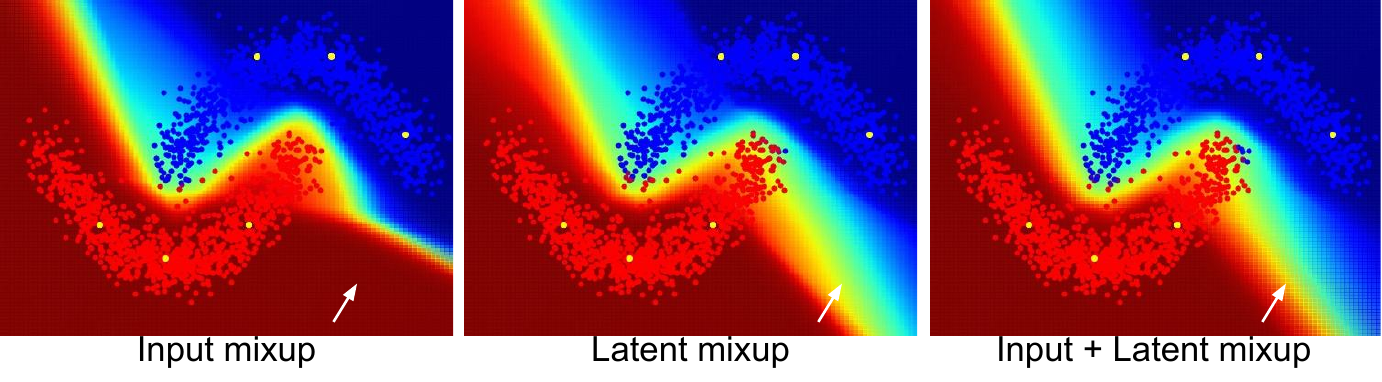}
\end{center}
\vspace{-0.3cm}
\caption{Decision boundary of SSL learning on two-moon toy data, where yellow dots represent the labeled data and the rest are unlabeled data.} 
\vspace{-0.3cm}
\label{fig:toy}
\end{figure}

Finally, we explore the effect of function smoothing 
brought by the presented SSL method. 
Starting with a two-moon toy data set, 
we observed in Fig. \ref{fig:toy} 
that mixing in the latent space 
increases the smoothness of the decision boundary 
in comparison to mixing at the input space only, 
an observation similar to \cite{verma2019manifold} for supervised learning. 
In addition, 
it also provided 
a broader range of uncertainty 
(broader region of low confidence) 
compared to mixing in input space only. 

\begin{figure}[t]
\begin{center}
\includegraphics[width=10cm,height=10cm,keepaspectratio]{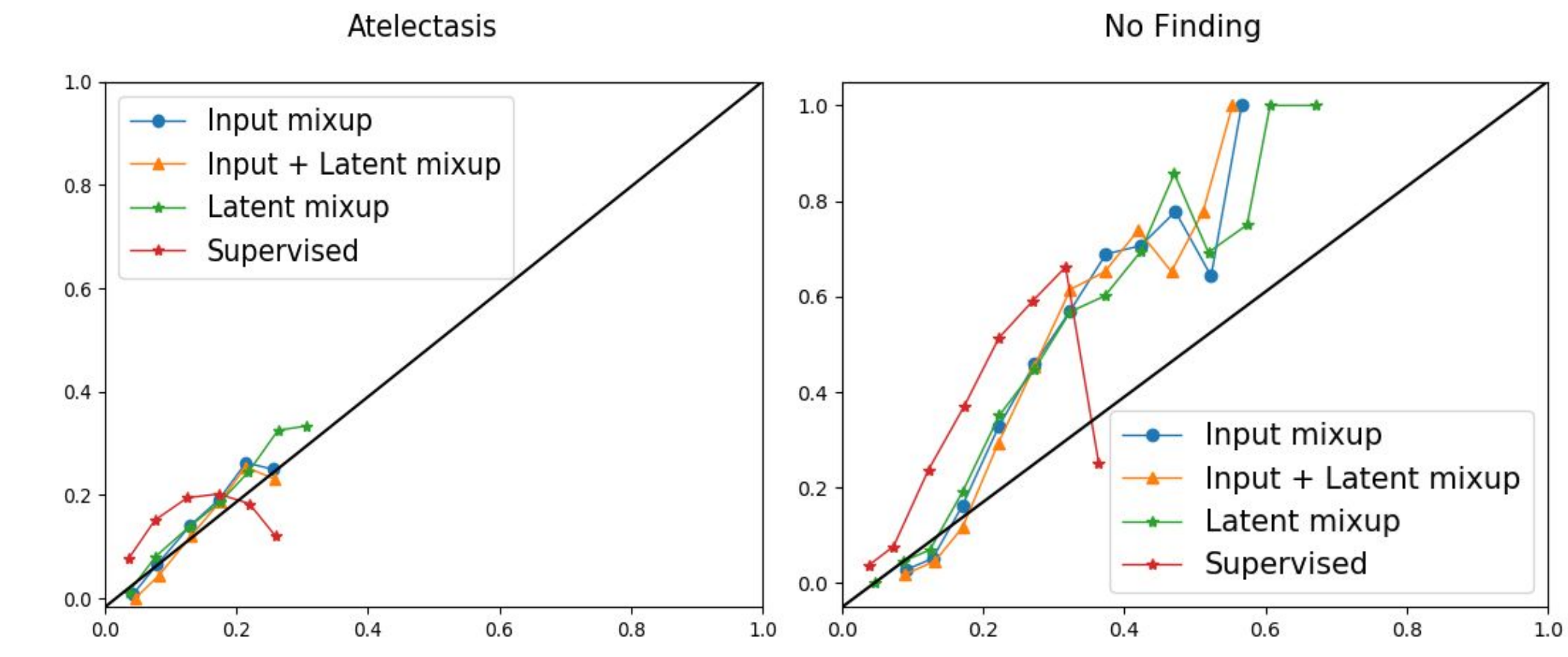}
\end{center}
    \vspace{-0.3cm}
    \caption{Reliability diagram of the networks on classifying two class labels from X-ray images, trained with $K$= 300 labeled data. Perfect calibration is indicated by the diagonal line representing identity function.} 
    \vspace{-0.3cm}
\label{fig:reliability}
\end{figure}

While it is not feasible to visualizing the decision boundary 
for the deep neural network in the presented medical image classification, 
we instead investigated the effect 
of a more smoothed confidence measure 
as observed in the toy data.
To do so, we 
consider the calibration of the model via the reliability diagram. 
Fig. \ref{fig:reliability} shows examples
of the network in classifying two class labels: 
as shown, 
in general, 
the mixup strategy improves the calibration of the network 
compared to a supervised baseline, 
while mixing at the latent space 
tends to further marginally improve the calibration 
compared to mixing at the input space alone. 

\section{Conclusion}
\vspace{-0.2cm}
We presented a novel semi-supervised learning method that 
regularizes the global smoothness of neural functions 
under the combination of input and latent mixing 
of labeled and unlabeled data. 
The evaluation on public chest X-ray data and skin disease data showed that the presented method 
improved the classification performance over 
SSL focusing on local smoothness of neural functions, 
as well as 
SSL regularizing global smoothness of the 
entire network between the input and output space. 
In future work, 
we are interested in extending the presented method for 
semi-supervised medical image segmentation. 


\noindent\textbf{Acknowledgement.} This work is supported by NSF CAREER ACI-1350374 and NIH NHLBI R15HL140500

\bibliographystyle{splncs04}
\bibliography{refs}
\end{document}